%% file: An_AlphaZero-Inspired_Approach_to_Solving_Search_Problems.tex
\documentclass{article}
\usepackage{amsthm,amssymb,amsmath,url,graphicx,algorithmic,color,tikz,authblk}
\theoremstyle{definition}

\input{definitions}
\begin{document}
\title{An AlphaZero-Inspired Approach to Solving Search Problems}
\author[1]{Evgeny Dantsin}
\author[2]{Vladik Kreinovich}
\author[1]{Alexander Wolpert}
\affil[1]{Roosevelt University, Chicago, U.S.}
\affil[2]{The University of Texas at El Paso, U.S.}

\date{}
\maketitle

\input{s0}

\input{s1}

\input{s2}
\input{s3}

\input{s4}

\input{s5}

\bibliographystyle{plain}
\bibliography{AI-refs}
\end{document}

%% file: definitions.tex
\newcommand{\sat}{\mathtt{SAT}}

\newcommand{\solver}{\mathcal{S}}
\newcommand{\easy}{\mathcal{E}}
\newcommand{\reduce}{\mathcal{R}}

\newcommand{\rl}{\mathcal{A}_1}
\newcommand{\dl}{\mathcal{A}_2}



%% file: s0.tex
\begin{abstract}
AlphaZero and its extension MuZero are computer programs that use machine-learning techniques to play at a superhuman level in chess, go, and a few other games. They achieved this level of play solely with reinforcement learning from self-play, without any domain knowledge except the game rules. It is a natural idea to adapt the methods and techniques used in AlphaZero for solving search problems such as the Boolean satisfiability problem (in its search version). Given a search problem, how to represent it for an AlphaZero-inspired solver? What are the ``rules of solving'' for this search problem? We describe possible representations in terms of \emph{easy-instance solvers} and \emph{self-reductions}, and we give examples of such representations for the satisfiability problem. We also describe a version of Monte Carlo tree search adapted for search problems.
\end{abstract}

%% file: s1.tex
\section{Introduction}


AlphaZero \cite{SHS+18} and its extension MuZero \cite{SAH+20} are computer programs developed by Google’s subsidiary DeepMind. They use machine-learning techniques to play at a superhuman level in chess, go, and a few other games. AlphaZero achieved this level of play solely with reinforcement learning from self-play, with no human data, no handcrafted evaluation functions, and no domain knowledge except the game rules. In comments on playing chess, the play style of AlphaZero is called ``alien'': it sometimes wins by making moves that would seem unthinkable to a human chess player.

The purpose of this paper is to adapt the methods and techniques used in AlphaZero for solving search problems such as, for example, the Boolean satisfiability problem (in its search version). Reinforcement learning has been applied to combinatorial optimization \cite{MSI+21,VSP+20}, but with using expert knowledge and handcrafted heuristics, which differs these applications from AlphaZero's approach. 

To design an AlphaZero-inspired solver for a search problem $\Pi$, we first need to represent $\Pi$ as a one-player combinatorial game, where the player starts from an initial position and tries to reach a winning position by making moves from one position to another. That is, we need to define what we mean by positions, winning positions, and possible moves.

We think of any instance of $\Pi$ as a possible position in the game. For example, if $\Pi$ is $\sat$, then any formula in CNF (a set of clauses) is viewed as a position. Certain instances of $\Pi$ are thought as ``easy'' instances, assuming that we already have an efficient solver for such instances. An easy instance plays the role of a winning position. In the case of $\sat$, a set of easy instances could contain formulas with the empty clause, formulas where each clause contains a pure literal, formulas in $2$-CNF, etc. By a possible move we mean a transition from an instance $x$ to an instance $x'$ such that the following holds:
\begin{itemize}
\item $x$ has a solution if and only if $x'$ has a solution;
\item a solution to $x$ can be computed from a solution to $x'$. 
\end{itemize}
The resolution rule gives examples of possible moves: $x'$ is obtained from $x$ by choosing two clauses and adding their resolvent to $x$. Another example is the pure literal elimination rule: $x'$ is obtained from $x$ by removing all clauses that contain pure literals. 

Thus, we define the ``game rules'' for $\Pi$ by specifying two components:
\begin{itemize}
\item a set of easy instances and a solver for this restriction of $\Pi$;
\item for each non-easy instance of $\Pi$, a set of possible moves from this instance. 
\end{itemize}
We call such a specification a \emph{setup} for solving $\Pi$. Section~\ref{sec:setup} gives a formal definition of setups in terms of \emph{easy-instance solvers} and \emph{self-reductions}. Section~\ref{sec:sat} gives examples of setups for $\sat$. 

Suppose we have chosen a setup for solving $\Pi$. A solver for $\Pi$ based on a given setup is described in Section~\ref{sec:solver}. This solver uses adapted versions of two key algorithms of AlphaZero: a \emph{reinforcement-learning algorithm} and a \emph{parameter-adjustment} algorithm. The former one uses Monte Carlo tree search to find a sequence of moves from an input instance of $\Pi$ to an easy instance. This algorithm has many parameters that are adjusted with help of the latter algorithm. The parameter-adjustment algorithm trains a deep neural network to find better values of the parameters; the choice of architecture of this network depends on how we represent instances of $\Pi$. 

The solver described in Section~\ref{sec:solver} can also be applied to another task called \emph{per-instance algorithm selection} \cite{HHL21,KHN+19}, In this task, we wish to design a ``meta-solver'' that solves a search problem $\Pi$ by automatically choosing (on a per-instance basis) a solver from a ``portfolio'' of solvers for $\Pi$, see Section~\ref{sec:end} for details. 


%% file: s2.tex
\section{Setups for Solving Search Problems}
\label{sec:setup}


\paragraph{Search problems.}
A \emph{search problem} is one of the standard types of computational problems. It is common to represent a search problem by a binary relation $R \subseteq X \times Y$ where $X$ is a set of \emph{instances} and $Y$ is a set of \emph{solutions}. If $(x,y) \in R$ then $y$ is called a \emph{solution to} $x$. 

The \emph{satisfiability problem} (in its search version) is an example of a search problem, see Section~\ref{sec:sat} for details. The corresponding set $X$ of instances consists of Boolean formulas in CNF. The set $Y$ of solutions consists of assignments of truth values to variables. An instance $x \in X$ has a solution $y \in Y$ if $y$ is a satisfying assignment of $x$. There are many ways to encode instances and solutions of the satisfiability problem; no particular encoding is specified in our example.


\paragraph{Solvers.}
Let $\Pi$ be a search problem. A \emph{solver} for $\Pi$ is an algorithm $\solver$ that either finds a solution, or reports that there is no solution, or may give up saying ``don't know''. That is, on every instance $x \in X$, 
\begin{itemize}
\item if $x$ has a solution, then $\solver$ returns some solution to $x$ or says ``don't know'';
\item if $x$ has no solution, then $\solver$ says ``no solution'' or says ``don't know''. 
\end{itemize}
Solvers may have parameters, additional input, and additional output. For example, in Section~\ref{sec:solver}, we describe a solver that takes as input not only an instance $x$ but also additional data $\theta$ with information about previous traces; the solver outputs an answer for $x$ and updates $\theta$.  


\paragraph{Easy instances.}
We assume that the set of instances of $\Pi$ has a designated subset $E \subseteq X$ whose elements are called \emph{easy} instances. The assumption behind $E$ is that it is ``easy''to determine whether an instance $x \in E$ has a solution and, moreover, if a solution exists, it is ``easy'' to find it. To formalize this assumption, we equip $\Pi$ with an algorithm denoted by $\easy$ and called an \emph{easy-instance solver}. On every instance $x \in X$, this algorithm determines whether $x$ is an easy instance and, if so, the algorithm finds a solution to $x$ or reports that $x$ has no solution:  
$$ 
\easy(x) = \left\{
\begin{array}{ll}
\mbox{``not easy''} & \mbox{if $x \notin E$} \\
\mbox{``no solution''} & \mbox{if $x \in E$ and $x$ has no solution} \\
\mbox{some solution to $x$} & \mbox{if $x \in E$ and $x$ has a solution}
\end{array}
\right.
$$
Section~\ref{sec:sat} gives examples of the set $E$ for the satisfiability problem. For example, $E$ can be the set of formulas $\phi$ such that $\phi$ is the empty set (this formula is satisfiable) or $\phi$ contains the empty clause (this formula is unsatisfiable). 


\paragraph{Self-reductions and moves.}
We define a \emph{self-reduction} of $\Pi$ to be a pair $r=(f_r,g_r)$, where $f_r$ and $g_r$ are computable functions such that for every instance $x \in X$,
\begin{itemize}
\item $f_r(x)$ is a finite set of instances;
\item if $x$ has a solution, then each instance in $f_r(x)$ has a solution;
\item for every instance $x' \in f_r(x)$ and for every solution $y \in Y$, if $y$ is a solution to $x'$, then $g_r(x, x',y)$ is a solution to $x$. 
\end{itemize}
If $x' \in f_r(x)$, then we say that the self-reduction $r$ offers a \emph{move} from $x$ to $x'$. Thus, for each instance, $f_r$ defines the set of all moves from this instance. We call $f_r$ the \emph{move function} of the self-reduction $r$. For a move from $x$ to $x'$, the function $g_r$ computes a solution to $x$ from a solution to $x'$ (if any). We call $g_r$ the \emph{solution function} of $r$. 

Examples of self-reductions of the satisfiability problem are given in Section~\ref{sec:sat}. Here we just mention two of them. The first example is a self-reduction $r=(f_r,g_r)$ where the move function $f_r$ is in fact the \emph{pure literal elimination} rule. This move function maps a CNF formula $\phi$ to a one-element set $\{\phi'\}$ where $\phi'$ is a CNF formula obtained from $\phi$ by successively removing all clauses containing pure literals. Another example is a self-reduction $r=(f_r,g_r)$ that uses the \emph{resolution rule}. For every CNF formula $\phi$, the set $f_r(\phi)$ consists of all CNF formulas obtained from $\phi$ by choosing two clauses and adding their resolvent to $\phi$. In both examples, the solution functions $g_r$ are defined in the obvious way, see Section~\ref{sec:sat} for details.


\paragraph{Paths.}
Let $\reduce$ be a finite set of self-reductions of $\Pi$. Let $x$ and $x'$ be instances of $\Pi$. By a \emph{path} from $x$ to $x'$ we mean a sequence
$$
x_0, r_1, x_1, r_2, x_2, \ldots, x_{n-1}, r_n, x_n
$$
where $x_0=x$, $x_n=x'$, and $r_i$ is a self-reduction from $\reduce$ that offers a move from $x_{i-1}$ to $x_i$ for all $i=1, \ldots, n$. Clearly, given such a path, we have the following:
\begin{itemize}
\item if $x$ has a solution, then $x'$ also has a solution;
\item if $y$ is a solution to $x'$, then $x$ has a solution that can be computed from $y$ by successively computing solutions to $x_{n-1}, \ldots, x_1, x_0$. 
\end{itemize}


\paragraph{Setups for solving.}
An easy-instance solver $\easy$ and a finite set $\mathcal{R}$ of self-reductions of $\Pi$ suggest the following approach to solving $\Pi$: 
\begin{enumerate}
\item Try to find a path from an input instance $x$ to an easy instance $x'$.
\item If such a path is found, either return a solution to $x$ (computed from a solution to $x'$) or return ``no solution'' (in the event that $x'$ has no solution). Otherwise, return ``don't know''.
\end{enumerate}
The key step here is a search for a path and its success depends on the choice of $\easy$ and $\mathcal{R}$. We call the pair $(\easy, \mathcal{R})$ a \emph{setup} for solving $\Pi$. Such a setup allows us to think of $\Pi$ as a one-player combinatorial game, where the player tries to find a sequence of moves from an initial position to a winning position. From this point of view, a setup for solving $\Pi$ defines the rules of this game. 

Note that, in general, a setup $(\easy, \mathcal{R})$ is not required to be ``complete'' in the following sense: for every instance $x$, there must be a path from $x$ to an easy instance. Section~\ref{sec:sat} shows examples of different setups for solving the satisfiability problem, including a setup where only satisfiable formulas have paths to easy instances. Solvers based on ``incomplete'' setups output ``don't know'' on instances that do not have paths to easy instances.   


%% file: s3.tex
\section{Examples for the Satisfiability Problem}
\label{sec:sat}

There are many possible setups for solving the satisfiability problem that make sense in our context. In this section, we give three setups for the purpose of illustration.

\paragraph{Satisfiability.}
Although the satisfiability problem is very well known and described in numerous books and articles \cite{BHM+21}, we give the basic definitions here to avoid ambiguity (notation and terminology slightly vary in the literature). 

Let $V=\{v_1, v_2, \ldots\}$ be a set of \emph{variables}. A \emph{literal} is a variable from $V$ or its negation; each of them is the \emph{complement} of the other. The complement of a literal $a$ is denoted by $\neg a$. A \emph{clause} is a finite set of literals that contains no pair of complements (a clause is thought of as the disjunction of its literals). A \emph{formula} is a finite set of clauses (a formula is thought of as the conjunction of its clauses). An \emph{assignment} is usually defined as a function from a finite set of variables to $\{0,1\}$, but it is convenient for us to use an equivalent definition: an \emph{assignment} is a finite set of literals without any pair of complements (this set is thought of as the conjunction of its literals). An assignment $\alpha$ \emph{satisfies} a clause $C$ if the intersection $\alpha \cap C$ is not empty. An assignment $\alpha$ \emph{satisfies} a formula $\phi$ if $\alpha$ satisfies every clause of $\phi$; we also call $\alpha$ a \emph{satisfying assignment} for $\phi$. 

The definitions above allow the empty clause and the empty formula. No assignment satisfies the empty clause and, thus, every formula with the empty clause is unsatisfiable. The formula consisting of only the empty clause is denoted by $\bot$. The empty formula (``no constraints at all'') is denoted by $\top$. By definition, $\top$ is satisfied by the empty assignment. 

It is common to denote the following decision problem by $\sat$: given a formula $\phi$, does it have a satisfying assignment? Slightly abusing this notation, we write $\sat$ to refer to the satisfiability problem in its search version: given a formula $\phi$, find a satisfying assignment or return ``no solution''. In terms of Section~\ref{sec:setup}, this search version is defined as follows. The set $X$ of instances consists of all formulas over $V$. The set $Y$ of solutions consists of all possible assignments, i.e., all finite subsets of literals over $V$ without any pair of complements. An assignment $\alpha \in Y$ is a solution to an instance $\phi \in X$ if and only if $\alpha$ satisfies $\phi$. 

\paragraph{Example 1: Setup based on resolutions.}
There are only two easy instances: $\top$ and $\bot$. Thus, an empty-instance solver $\easy$ is trivial. A set $\mathcal{R}$ of self-reductions consists of the following three self-reductions commonly used in $\sat$ solving:
\begin{itemize}
\item \textit{Resolution rule.} Let $\phi$ be a formula with clauses $C_1$ and $C_2$ such that $C_1$ contains a literal $a$ and $C_2$ contains its complement $\neg a$. If the set $C_1 \cup C_2 - \{a, \neg a\}$ contains no pair of complements, then we call this set the \emph{resolvent} of $C_1$ and $C_2$. If $\phi'$ is the formula obtained from $\phi$ by adding this resolvent, we say that $\phi'$ is obtained from $\phi$ by the \emph{resolution rule}. The \emph{resolution self-reduction} is a pair $r=(f_r,g_r)$ where the move function is defined by
$$
f_r(\phi) = \{\phi' \ | \ \mbox{$\phi'$ is obtained from $\phi$ by the resolution rule}\}
$$
and the solution function $g_r$ is defined as follows: for all formulas $\phi$, if $\alpha$ is a solution to a formula $\phi' \in f_r(\phi)$ then $g_r(\phi,\phi',\alpha)$ is $\alpha$. 
\item \textit{Subsumption rule.} If a clause $C_1$ is a proper subset of a clause $C_2$, we say that $C_1$ is \emph{subsumed} by $C_2$ and we call the clause $C_2$ \emph{unnecessary}. The \emph{subsumption self-reduction} is the following self-reduction $r=(f_r,g_r)$. The move function $f_r$ maps a formula $\phi$ to a one-element set $\{\phi'\}$ where the formula $\phi'$ is obtained from $\phi$ by removing all unnecessary clauses. The solution function $g_r$ is obvious: $g_r(\phi,\phi',\alpha) = \alpha$ for all $\phi$, $\phi'$, and $\alpha$.
\item \textit{Pure literal elimination.} A literal $a$ in $\phi$ is called a \emph{pure} literal if no clause of $\phi$ contains $\neg a$. The \emph{pure literal self-reduction} $r=(f_r,g_r)$ is defined as follows. The move function $f_r$ maps a formula $\phi$ to a one-element set $\{\phi'\}$ where $\phi'$ is obtained from $\phi$ by successively removing all clauses containing pure literals until $\phi'$ has no pure literals. If $\alpha$ is a satisfying assignment for $\phi'$, then $g_r(\phi,\phi',\alpha)$ is the extension of $\alpha$ that assigns ``true'' to all pure literals in $\phi$.
\end{itemize}
This setup $(\easy,\mathcal{R})$ is ``complete'': for every formula $\phi$, there is a path from $\phi$ to either $\top$ or $\bot$, see for example \cite{BN21}.

\paragraph{Example 2: Setup based on resolutions and the extension rule.}
The setup described above can be extended by adding a self-reduction based on the \emph{extension rule} \cite{Tse68}. Let $\phi$ be a formula and let $v$ be a variable not appearing in $\phi$: no clause of $\phi$ contains $v$ or $\neg v$. Let $a$ and $b$ be literals such that their underlying variables appear in $\phi$. The extension rule adds clauses 
$$
\{a, \neg v\}, \ \{b, \neg v\}, \ \{\neg a, \neg b, v\}
$$
to $\phi$. In the corresponding self-reduction $r=(f_r,g_r)$, the move function $f_r$ is defined by
$$
f_r(\phi) = \{\phi' \ | \ \mbox{$\phi'$ is obtained from $\phi$ by the extension rule}\}
$$
and the solution function $g_r$ is the same as in the resolution self-reduction: $g_r(\phi,\phi',\alpha)$ is $\alpha$. 

The extension rule makes resolution proof systems much stronger, but there are no good heuristics for choosing extension literals $a$ and $b$. This problem of using the extension rule in practical $\sat$ solvers is discussed in \cite[section 7.8]{BN21}, where the authors note that ``if this could be done well, the gains would be enormous'' and ``the main bottleneck appears to be that we have no good heuristics for how to choose extension formulas''.

\paragraph{Example 3: Setup based on flipping.}
A variable is called a \emph{positive} literal; its negation is called a \emph{negative} literal. We define an easy instance to be a formula in which every clause has at least one positive literal. Obviously, such a formula is satisfied by the set of these positive literals. An algorithm $\easy$ determines whether an instance $\phi$ is an easy instance and if so, $\easy(\phi)$ is the corresponding set of positive literals. The \emph{flipping rule} transforms a formula $\phi$ taking the following two steps:
\begin{enumerate}
\item Choose a clause $C \in \phi$ in which all literals are negative (if $\phi$ is not an easy instance, such a clause exists).
\item Choose a literal $\neg v_i \in C$ and ``flip'' all of its occurrences in $\phi$, i.e., replace $\neg v_i$ with $v_i$ everywhere in $\phi$. 
\end{enumerate} 
We can define $\mathcal{R}$ to be a set of one or more self-reductions based on the \emph{flipping} rule. The move function in a such a self-reduction maps $\phi$ into a set of formulas obtained from $\phi$ by applying the flipping rule. Note that the setup $(\easy,\mathcal{R})$ is not ``complete''. If $\phi$ is satisfiable, then there is a path from $\phi$ to an easy instance. Otherwise, $\phi$ has no path to any easy instance (all easy instances are satisfiable).

%% file: s4.tex
\section{Solvers Based on Setups}
\label{sec:solver}


Let $(\easy,\reduce)$ be a setup for solving a search problem $\Pi$. We describe a solver for $\Pi$ based on this setup. This solver, denoted by $\solver$, tries to find a path from an input instance $x$ to an easy instance and, if such a path is found, $\solver$ outputs an answer for $x$. The solver has parameters whose values change from run to run, and $\solver$ updates these values itself. The key point is that $\solver$ uses machine-learning techniques for both tasks, namely, for a path search and for updating values of the parameters. Roughly, $\solver$ uses a reinforcement-learning algorithm $\rl$ to search for a path and it uses a parameter-adjustment algorithm $\dl$ to search for ``better'' values of the parameters. We first describe a bird's eye view of $\solver$ and then give more details. 

\paragraph{Input and output.}
The input to $\solver$ has two parts: and instance $x \in X$ and a binary string $\theta \in \{0,1\}^*$ called a \emph{parameter string}. This string encodes values of the parameters of $\solver$ and information about the solver's previous traces. We assume that $\theta$ is stored in a data store outside $\solver$. We also assume that $\theta$ is initialized before the first run of $\solver$ and it is updated after each next run. Thus, the output of $\solver$ on $x$ and $\theta$ is an answer for $x$ (either a solution to $x$, or ``no solution'', or ``don't know'') and the updated parameter string $\theta'$. 

\paragraph{Solver $\solver$.}
On input $x$ and $\theta$, the solver $\solver$ works as follows:
\begin{enumerate}
\item Run $\rl$. This algorithm produces a path 
\begin{equation}
\label{eq:path}
x_0, r_1, x_1, r_2, x_2, \ldots, x_{n-1}, r_n, x_n
\end{equation}
where $x_0=x$. Note that $x_n$ is not necessarily an easy instance. In the course of producing this path, $\rl$ generates other paths and measures the ``quality'' of the moves occurring in these paths: one move is better than another if it is expected to have a better chance of leading to an easy instance. Information about the quality of the moves is stored as \emph{quality data} $\delta$. The output of $\rl$ is path (\ref{eq:path}) and $\delta$.
\item Return an answer for $x$:
\begin{enumerate}
\item If $\easy(x_n)$ is ``not easy'', then return ``don't know''.
\item If $\easy(x_n)$ is ``no solution'', then return ``no solution''.
\item If $\easy(x_n)$ is a solution to $x_n$, work backwards from $x_n$ and use the solution functions $g_r$ from $\reduce$ to find successively solutions to $x_{n-1}, \ldots, x_0$. Finally, return the solution to $x_0$.
\end{enumerate}
\item Run $\dl$. This algorithm takes $\delta$ and merges it with similar quality data collected in the previous runs of $\solver$. The result of merge is used for training and updating parameters $\theta$ to new parameters $\theta'$.
\item Return the updated parameter string $\theta'$ for storing.
\end{enumerate}


\paragraph{Reinforcement-learning algorithm $\rl$.}
This algorithm is a Monte Carlo tree search algorithm adapted for search problems. More exactly, $\rl$ is a version of the Adaptive Multistage Sampling algorithm (AMS) described in \cite{CFH+05}. The algorithm $\rl$ cannot apply AMS as a black-box algorithm because the input to AMS is not given explicitly. Instead, $\rl$ supplies the input data in a ``just-in-time'' manner as follows.
\begin{itemize}
\item \textit{Initialization of rewards.} 
In each recursive call, AMS initializes rewards of moves. Given an instance $x$, the \emph{reward} of a move is a measure for the belief that this move is on a bounded-length path from $x$ to an easy instance. The reward is maximum if the move is on such a path to an easy instance. The algorithm $\rl$ needs a belief estimation algorithm that computes initial reward values for moves. This estimation is implemented by a deep neural network $N$ that uses parameters given in $\theta$. The initial rewards are improved by training this network. 
\item \textit{Sampling algorithm.} 
The algorithm $\rl$ provides a \emph{sampling algorithm} for AMS. On an instance $x$, this algorithm uses the parameters in $\theta$ to sample the moves from $f_r(x)$ for each self-reduction $r \in \reduce$.  The sampling algorithm can be  implemented using the same deep neural network $N$, or it can be a different neural network that shares weights with $N$. The distributions for self-reductions are improved by training $N$, which means that the improved distributions assign higher probabilities to moves with higher accumulated rewards.  
\item \textit{Output.}
According to the description of AMS in \cite{CFH+05}, this algorithm returns a path that has the maximum accumulated reward. In addition to this optimal path, $\rl$ collect the following quality data $\delta$ and returns it for training:
\begin{itemize} 
\item for every instance $x$ and every self-reduction $r$ explored in the run, the accumulated probability distribution on $f_r(x)$;
\item for every instance $x$ explored in the run, the accumulated quality of $x$ (``value'' of $x$ in the AMS terminology).
\end{itemize}
\end{itemize}


\paragraph{Parameter-adjustment algorithm $\dl$.}
After taking the quality data $\delta$ and merging it with similar datasets, $\dl$ trains the deep neural network $N$ to adjust the parameters $\theta$. Note that the choice of architecture of $N$ is dictated by instance representation. For example, convolutional neural networks can be used in the case of finite dimensional tensor representation. If instances are represented by binary strings of variable length, recurrent neural networks can be used. In the case of $\sat$, it is natural to represent instances by graphs and, therefore, $N$ can be implemented as a graph neural network. In particular an extension of the network constructed in \cite{SLB+19} could be used. Also note that the architecture of $N$ determines what instance features can be discovered from training. 

What datasets can be used for the initial training of $N$? It is more or less common to train a neural network using randomly shuffled data. Certain sets of instances (for example, industrial instances of $\sat$) expose self-similarity: large instances have the same properties as smaller ones. In such cases, it makes sense to train $N$ using \emph{curriculum learning} \cite{NPL+20} where the training starts from samples of small size and moves to larger ones.


%% file: s5.tex
\section{Concluding Remark}
\label{sec:end}


In this paper, we described how to adapt AlphaZero's techniques for designing a solver for a search problem. This adaptation can also be used for another task called \emph{per-instance algorithm selection} \cite{HHL21,KHN+19}. In this task, we are given a search problem $\Pi$ and a ``portfolio'' of solvers for $\Pi$. We wish to design a ``meta-solver'' that automatically chooses a solver from the portfolio on a per-instance basis and, thereby, it achieves better performance than any single solver from the portfolio.

Suppose all solvers in the portfolio are of the following type. Such a solver takes as input an instance $x$ of $\Pi$ and produces another instance $x'$ such that (1) $x'$ has a solution if and only if $x$ has a solution and (2) a solution to $x$ can be computed from a solution to $x'$. If $x'$ is an easy instance then the solver returns an answer, otherwise the solver returns $x'$ and says``don't know''. Many $\sat$ solvers are of this type, for example, iterative solvers like resolution-based solvers with a limited number of iterations. The portfolio with such solvers can be viewed as a self-reduction where the move function maps an input instance $x$ to the set of all instances $x'$ produced by the solvers. Thus, we can use the solver described in Section~\ref{sec:solver} as a meta-solver for $\Pi$.